\documentclass[a4,a4wide]{article}
\usepackage{aaai}
\usepackage{times}
\usepackage{helvet}
\usepackage{courier}
\usepackage{graphicx}
\usepackage{caption}
\usepackage{subcaption}

\newcommand{\tbeg}{\langle}
\newcommand{\tend}{\rangle}

\newcommand{\lpnot}{\mbox{not}\;\,}
\newcommand{\lpor}{\;\,\mbox{\textsc{or}}\;\,}

\newcommand{\hif}{\leftarrow}

\newcommand{\aspindent}{\hspace*{.5in}}
\title{An ASP-Based Architecture for Autonomous UAVs in Dynamic Environments: Progress Report}
\author{Marcello Balduccini \and William C.~Regli \and Duc N.~Nguyen\\
Applied Informatics Group\\
Drexel University\\
Philadelphia, PA, USA
}

\begin{document}

\nocopyright

\maketitle
\begin{abstract}
\begin{quote}

Traditional AI reasoning techniques have been used successfully in many domains, including logistics, scheduling and game playing.  This paper is part of a project aimed at investigating how such techniques can be extended to coordinate teams of unmanned aerial vehicles (UAVs) in dynamic environments.  Specifically challenging are real-world environments where UAVs and other network-enabled devices must communicate to coordinate---and communication actions are neither reliable nor free.  Such network-centric environments are common in military, public safety and commercial applications, yet most research (even multi-agent planning) usually takes communications among distributed agents as a given. We address this challenge by developing an agent architecture and reasoning algorithms based on Answer Set Programming (ASP). ASP has been chosen for this task because it enables high flexibility of representation, both of knowledge and of reasoning tasks. Although ASP\ has been used successfully in a number of applications, and ASP-based architectures have been studied for about a decade, to the best of our knowledge this is the first practical application of a complete ASP-based agent architecture. It is also the first practical application of ASP involving a combination of centralized reasoning, decentralized reasoning, execution monitoring, and reasoning about network communications. This work has been empirically validated using a distributed network-centric software evaluation testbed and the results provide guidance to designers in how to understand and control intelligent systems that operate in these environments.

\end{quote}
\end{abstract}

\section{Introduction}
Unmanned Aerial Vehicles (UAVs) promise to revolutionize the way in which we use our airspace.  From talk of automating the navigation for major shipping companies to the use of small helicopters as "deliverymen" that drop your packages at the door, it is clear that our airspaces will become increasingly crowded in the near future.  This increased utilization and congestion has created the need for new and different methods of coordinating assets using the airspace.  Currently, airspace management is the job for mostly human controllers. As the number of entities using the airspace vastly increases---many of which are autonomous---the need for improved autonomy techniques becomes evident. 

The challenge in an environment full of UAVs is that the world is highly dynamic and the communications environment is uncertain, making coordination difficult. Communicative actions in such setting are neither reliable nor free. 

\begin{figure*}[htbp]
  \begin{subfigure}[t]{.45\linewidth}
 \centering
 \includegraphics[width=\columnwidth]{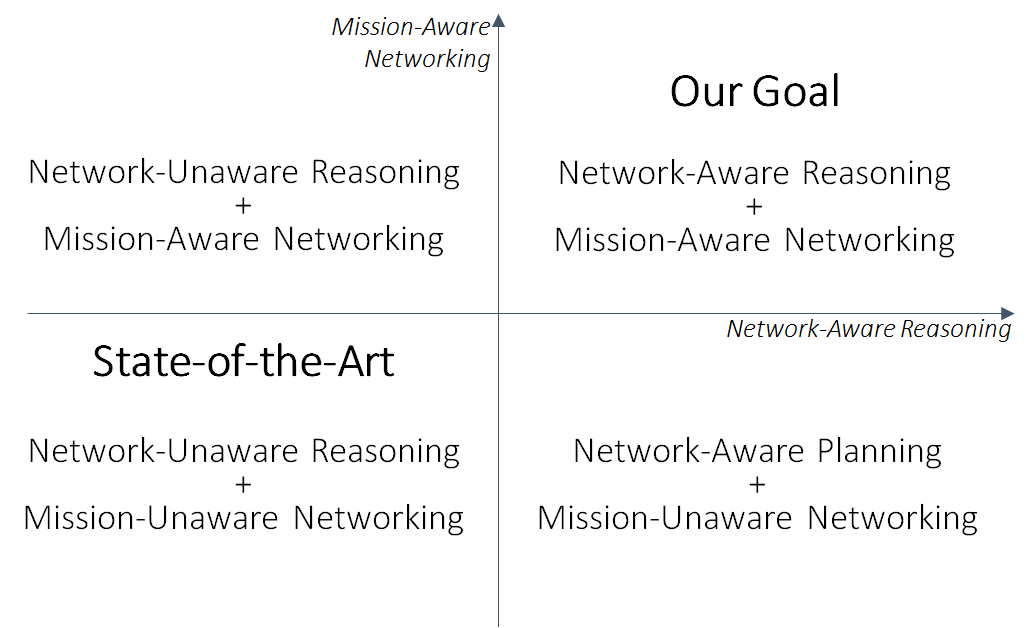}
% \subcaption{The current state of reasoning and networking (lower-left) vs our goal combination (top-right)}
  \end{subfigure}
  \quad\quad
  \begin{subfigure}[t]{.45\linewidth}
    \centering
        \includegraphics[width=\columnwidth]{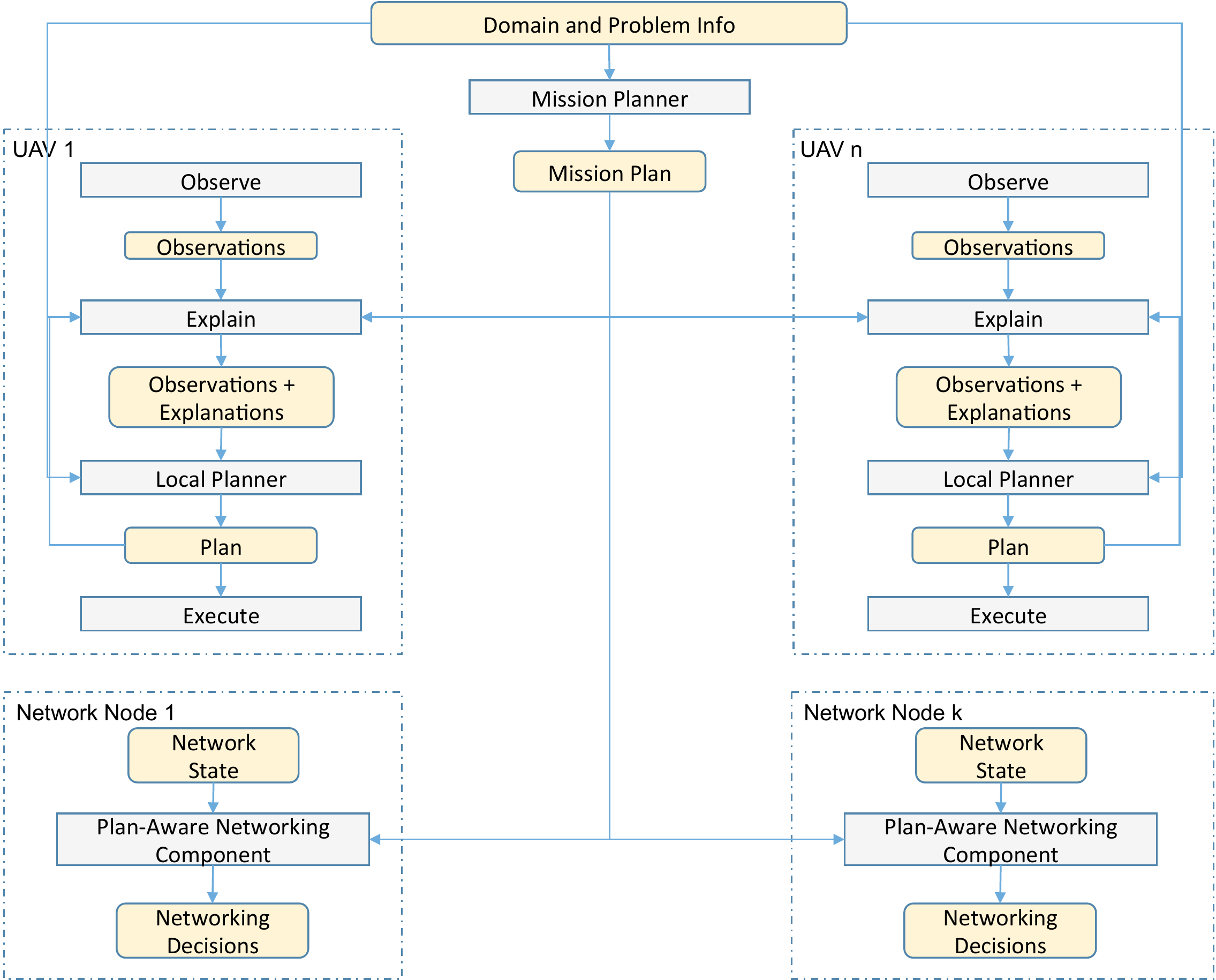}
%    \subcaption{(a) Information flow in our framework.}
  \end{subfigure}
\caption{(a) The current state of reasoning and networking (lower-left) vs our goal combination (top-right); (b) Information flow in our framework.}
\label{fig:quadchart}
\label{fig:arch}
\end{figure*}

The work discussed here is in the context of the development of a novel application of network-aware reasoning and of an intelligent mission-aware network layer to the problem of UAV coordination. Typically, AI reasoning techniques do not consider realistic network models, nor does the network layer reason dynamically about the needs of the mission plan. With network-aware reasoning (Figure~\ref{fig:quadchart}a), a reasoner (either centralized or decentralized) factors in the communications network and its conditions, while with mission-aware networking, an intelligent network middleware service considers the mission and network state, and dynamically infers quality of service (QoS) requirements for mission execution.

In this paper we provide a general overview of the approach, and then focus on the aspect of network-aware reasoning. We address this challenge by developing an agent architecture and reasoning algorithms based on Answer Set Programming (ASP, \cite{gl91,mt99,bar03}). ASP has been chosen for this task because it enables high flexibility of representation, both of knowledge and of reasoning tasks. Although ASP\ has been used successfully in a number of applications, and ASP-based architectures have been studied for about a decade, to the best of our knowledge this is the first practical application of a complete ASP-based agent architecture. It is also the first practical application of ASP\ involving a combination of centralized reasoning, decentralized reasoning, execution monitoring, and reasoning about network communications. This work has been empirically validated using a distributed network-centric software evaluation testbed and the results provide guidance to designers in how to understand and control intelligent systems that operate in these environments.

The next section describes relevant systems and reasoning techniques, and is followed by a motivating scenario that applies to UAV coordination. The Technical Approach section describes network-aware reasoning and demonstrates the level of sophistication of the behavior exhibited by the UAVs using example problem instances. Next is a description of the network-centric evaluation testbed used for simulations. Finally, we draw conclusions and discuss future work.

\section{Related Work}\label{sec:background}
Incorporating network properties into planning and decision-making has been investigated in~\cite{usbeckCR12}. The authors' results indicate that plan execution effectiveness and performance is increased with the increased network-awareness during the planning phase. The UAV coordination approach in this current work combines network-awareness during the reasoning processes with a plan-aware network layer. 

The problem of mission planning for UAVs under communication constraints has been addressed in~\cite{kpj13}, where an ad-hoc task allocation process is employed to engage under-utilized UAVs as communication relays. In our work, we do not separate planning from the engagement of under-utilized UAVs, and do not rely on ad-hoc, hard-wired behaviors. Our approach gives the planner more flexibility and finer-grained control of the actions that occur in the plans, and allows for the emergence of sophisticated behaviors without the need to pre-specify them.

The architecture adopted in this work is an evolution of \cite{bg08}, which can be viewed as an instantiation of the BDI agent model \cite{rg91,woo00}. Here, the architecture has been extended to include a centralized mission planning phase, and to reason about other agents' behavior. Recent related work on logical theories of intentions \cite{bgb14} can be further integrated into our approach to allow for a more systematic hierarchical characterization of  actions, which is likely to increase performance.   

Traditionally, AI planning techniques have been used (to great success) to perform  multi-agent teaming, and UAV coordination. Multi-agent teamwork decision frameworks such as the ones described in \cite{pt02} may factor communication costs into the decision-making. However, the agents do not actively reason about other agent's observed behavior, nor about the communication process. Moreover, policies are used as opposed to reasoning from models of domains and of agent behavior. 

The reasoning techniques used in the present work have already been successfully applied to domains ranging from complex cyber-physical systems to workforce scheduling. To the best of our knowledge, however, they have never been applied to domains combining realistic communications and multiple agents.   

Finally, high-fidelity multi-agent simulators (e.g., AgentFly~\cite{sislak2012agentfly}) do not account for network dynamism nor provide a realistic network model. For this reason, we base our simulator on the Common Open Research Emulator (CORE)~\cite{ahrenholz10core}. CORE provides network models in which communications are neither reliable nor free.

\section{Motivating Scenario}\label{sec:scenario}
To motivate the need for network-aware reasoning and mission-aware networking, consider a simple UAV coordination problem, depicted in Figure~\ref{subfig:datamule}, in which two UAVs are tasked with taking pictures of a set of three targets, and with relaying the information to a home base.

%\begin{figure}[t]
% \centering%\begin{centering}
% \includegraphics[width=.85\columnwidth]{figs/screens/inst2_00_crop.png}
% %\end{centering}
% \caption{An example problem instance for UAV coordination. The home base is the black node in the lower left corner and the targets are shown as red dots in the upper right corner. Relays form a mesh and extend the network.}\label{fig:scenario}
%\end{figure}

Fixed relay access points extend the communications range of the home base. The UAVs can share images of the targets with each other and with the relays when they are within radio range. The simplest solution to this problem consists in entirely disregarding the networking component of the scenario, and generating a mission plan in which each UAV flies to a different set of targets, takes pictures of them, and flies back to the home base, where the pictures are transferred. This solution, however, is not satisfactory. First of all, it is inefficient, because it requires that the UAVs fly all the way back to the home base before the images can be used. The time it takes for the UAVs to fly back may easily render the images too outdated to be useful. 
Secondly, disregarding the  network during the reasoning process may lead to mission failure --- especially in the case of unexpected events, such as enemy forces blocking transit to and from the home base after a UAV has reached a target. Even if the UAVs are capable of autonomous behavior, they will not be able to complete the mission unless they take advantage of the network.

Another common solution consists of acknowledging the availability of the network, and assuming that the network is constantly available throughout plan execution. A corresponding mission plan would instruct each UAV to fly to a different set of targets, and take pictures of them, while the network relays the data back to the home base. This solution is \emph{optimistic} in that it assumes that the radio range is sufficient to reach the area where the targets are located, and that the relays will work correctly throughout the execution of the mission plan.

This optimistic solution is more efficient than the previous one, since the pictures are received by the home base soon after they are taken. Under realistic conditions, however, the strong assumptions it relies upon may easily lead to mission failure---for example, if the radio range does not reach the area where the targets are located. 

 In this work, the reasoning processes take into account not only the presence of the network, but also its configuration and characteristics, taking advantage of available resources whenever possible. The mission planner is given information about the radio range of the relays and determines, for example, that the targets are out of range. A possible mission plan constructed by this information into account consists in having one UAV fly to the targets and take pictures, while the other UAV remains in a position to act as a network bridge between the relays and the UAV that is taking pictures. This solution is as efficient as the optimistic solution presented earlier, but is more robust, because it does not rely on the same strong assumptions.  

Conversely, when given a mission plan, an intelligent network middleware service capable of sensing conditions and modifying network parameters (e.g., modify network routes, limit bandwidth to certain applications, and prioritize network traffic) is able to adapt the network to provide optimal communications needed during plan execution. A relay or UAV running such a middleware is able to interrupt or limit bandwidth given to other applications to allow the other UAV to transfer images and information toward home base. Without this traffic prioritization, network capacity could be reached prohibiting image  transfer.

%In this paper, we focus on an episodic version of this problem, and on
%the problem of creating mission plans that allow the UAVs to most
%effectively travel to a collection of targets and return information
%about them to a home base.

\section{Technical Approach}
In this section, we formulate the problem in more details; provide technical background; discuss the design of the agent architecture and of the reasoning modules; and demonstrate the sophistication of the resulting behavior of the agents in two scenarios.

\subsection{Problem Formulation}
%\emph{put problem formulation in from the slides}
%slide 2 and 8

A problem instance for coordinating UAVs to observe targets and deliver information (e.g., images) to a home base is defined by a set of UAVs, $u_1, u_2, \ldots$, a set of targets, $t_1, t_2, \ldots$, a (possibly empty) set of fixed radio relays, $r_1, r_2, \ldots$, and a home base. The UAVs, the relays, and the home base are called radio nodes (or network nodes). Two nodes are in  radio contact if they are within a distance $\rho$ from each other, called radio range\footnote{For simplicity, we assume that all the radio nodes use comparable network devices, and that thus $\rho$ is unique throughout the environment.}, or if they can relay information to each other through intermediary radio nodes that are themselves within radio range. The UAVs are expected to travel from the home base to the targets to take pictures of the targets and deliver them to the home base. A UAV will automatically take a picture when it reaches a target. If a UAV is within radio range of a radio node, the pictures are automatically shared.
From the UAVs' perspective, the environment is only partially observable. Features of the domain that are observable to a UAV $u$ are (1) which radio nodes $u$ can and cannot communicate with by means of the network, and (2) the position of any UAV that near $u$. 

The goal is to have the UAVs take a picture of each of the targets so that (1) the task is accomplished as quickly as possible, and (2) the total ``staleness'' of the pictures is as small as possible. Staleness is defined as the time elapsed from the moment a picture is taken, to the moment it is received by the home base. While the UAVs carry on their tasks, the relays are expected to actively prioritize traffic over the network in order to ensure mission success and further reduce staleness.%It is also important to note that a picture is only taken when the target is visited for the first time for planner simplicity, however, it is not difficult to extend to allow multiple pictures to be taken of a target to maintain fresh pictures.

\subsection{Answer Set Programming}

In this section we provide a definition of the syntax of ASP and of its informal semantics. We refer the 
reader to \cite{gl91,ns00,bar03} for a specification of the formal semantics.
Let $\Sigma$ be a signature containing constant, function and
predicate symbols. Terms and atoms are formed as usual in first-order logic. A (basic)
literal is either an atom $a$ or its strong (also called classical or epistemic) negation
$\neg a$. 
%The sets of atoms and literals formed from $\Sigma$ are denoted
%by $atoms(\Sigma)$ and $literals(\Sigma)$ respectively.
A \emph{rule} is a statement of the form:
%\begin{equation}\label{eq:rule}
\[
h_1 \lpor \ldots \lpor h_k \hif l_1, \ldots, l_m, \lpnot l_{m+1}, \ldots, \lpnot l_n
\]
%\end{equation}
where $h_i$'s and $l_i$'s are ground literals and $\mbox{\emph{not}}$ is the so-called
\emph{default negation}. The intuitive meaning of the rule is that a
reasoner who believes $\{ l_1, \ldots, l_m \}$ and has no reason to believe
$\{l_{m+1}, \ldots, l_n\}$, must believe one of $h_i$'s.
Symbol $\hif$ can be omitted if no $l_i$'s are specified.
%We call $\{ h_1, \ldots, h_k \}$ the \emph{head} of the rule, and
%$\{ l_1, \ldots, l_m, \lpnot l_{m+1}, \ldots, \lpnot l_n \}$ the \emph{body}
%of the rule. 
%Given a rule $r$, we denote its head and body by $head(r)$ and
%$body(r)$ respectively.
%
%\st
Often, rules of the form
$h \hif \lpnot h, l_1, \ldots, \lpnot l_n$ are abbreviated
into $\hif l_1, \ldots, \lpnot l_n$, and called \emph{constraints}.
The intuitive meaning of a constraint is that
$\{ l_1, \ldots, l_m, \lpnot l_{m+1}, \ldots, \lpnot l_n \}$ must
not be satisfied. A rule containing variables is interpreted as
the shorthand for the set of rules obtained by replacing the variables
with all the possible ground terms.
%
%\st
A \emph{program} is a pair $\tbeg \Sigma, \Pi \tend$, where $\Sigma$ is
a signature and $\Pi$ is a set of rules over $\Sigma$. We often
denote programs just by the second element of the pair, and let the
signature be defined implicitly.
%In that case, the signature of $\Pi$ is denoted by $\Sigma(\Pi)$.
Finally, the \emph{answer set} (or \emph{model}) of a program $\Pi$ is 
the collection of its consequences under the answer set semantics.
Notice that the semantics of ASP is defined in such a way that programs
may have multiple answer sets, intuitively corresponding
to alternative solutions satisfying the specification given by the program.
The semantics of default negation provides a simple
way of encoding choices. For example, the set of rules
$\{ p \hif \lpnot q.\ \ q \hif \lpnot p. \}$ intuitively states that
either $p$ or $q$ may hold, and the corresponding program has two answer
sets, $\{ p \}$, $\{ q \}$.
The language of ASP has been
extended with \emph{constraint literals} \cite{ns00}, which are
expressions of the form $m \{ l_1, l_2, \ldots, l_k \} n$,
where $m$, $n$ are arithmetic expressions and $l_i$'s are basic literals
as defined above. A constraint literal is satisfied whenever the
number of literals that hold from $\{l_1, \ldots, l_k\}$ is between $m$ and $n$,
inclusive. Using constraint literals, the choice between $p$ and $q$,
under some set of conditions $\Gamma$,
can be compactly encoded by the rule $1 \{ p, q \} 1 \hif \Gamma$.
A rule of this kind is called \emph{choice rule}.
To further increase flexibility, the set $\{ l_1, \ldots, l_k \}$ can
also be specified as $\{ l(\vec{X}) : d(\vec{X}) \}$, where $\vec{X}$ is a list
of variables. Such an expression intuitively stands for the set of all $l(\vec{x})$
such that $d(\vec{x})$ holds. We refer the reader
to \cite{ns00} for a more detailed definition of the syntax of constraint
literals and of the corresponding extended rules.

\subsection{Agent Architecture}

The architecture used in this project follows the BDI agent model \cite{rg91,woo00}, which provides a good foundation because of its logical underpinning, clear structure and flexibility. In particular, we build upon ASP-based instances of this model \cite{bg00,bg08} because they employ directly-executable logical languages  featuring good computational properties while at the same time ensuring elaboration tolerance \cite{mcc98} and elegant handling of incomplete information, non-monotonicity, and dynamic domains.

A sketch of the information flow throughout the system is shown in Figure~\ref{fig:arch}b.\footnote{The tasks in the various boxes are executed only when necessary.} Initially, a centralized \emph{mission planner} is given a description of the domain and of the problem instance, and finds a plan that uses the available UAVs to achieve the  goal.

\if 0
\begin{figure}[t]
        \includegraphics[width=.6\columnwidth]{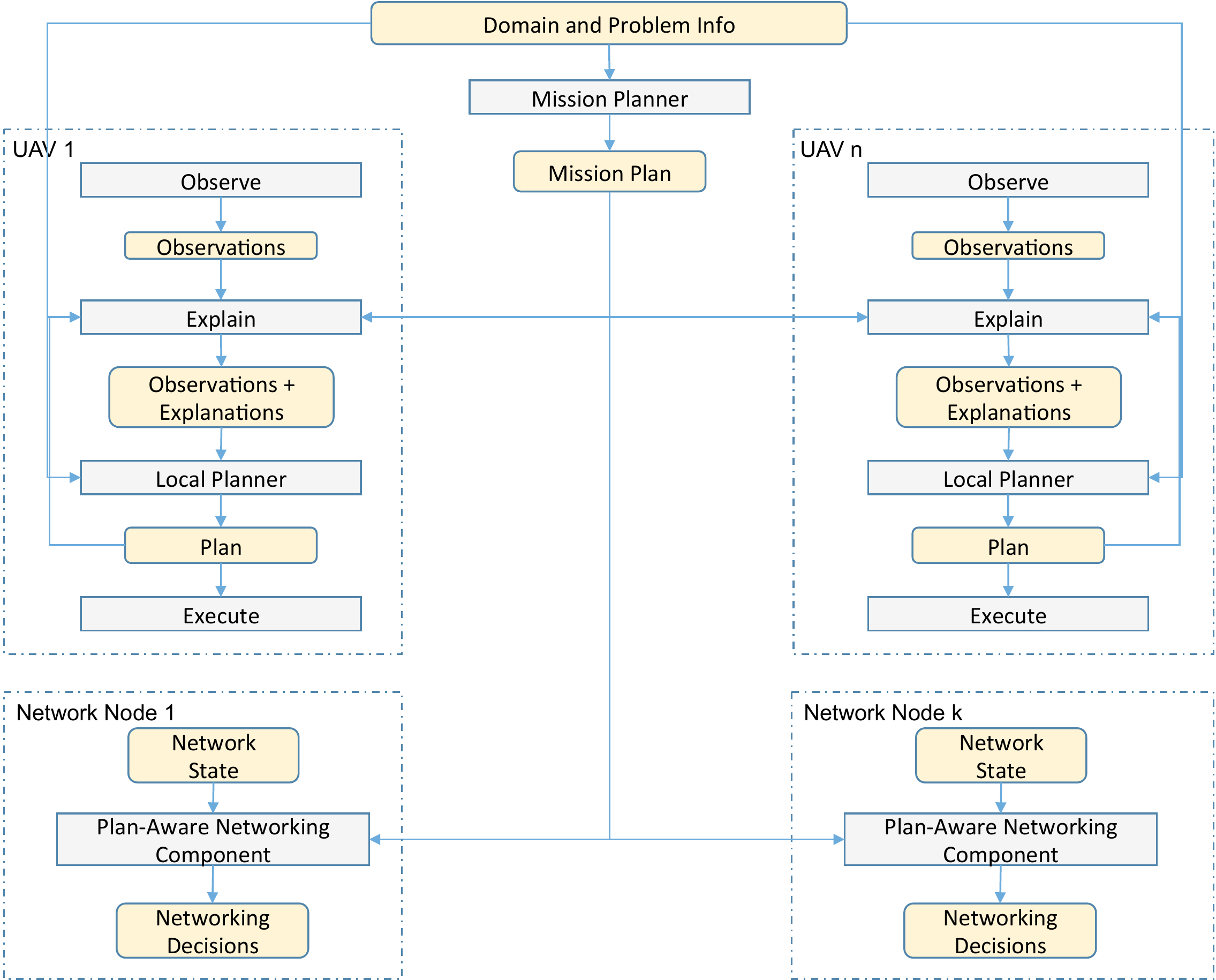}
\caption{Information flow in our framework.}\label{fig:arch}
\end{figure}
\fi
Next, each UAV receives the plan and begins executing it individually.
As plan execution unfolds, the communication state changes, potentially affecting network connectivity. For example, the UAVs may move in and out of range of each other and of the other network nodes. Unexpected events, such as relays failing or temporarily becoming disconnected, may also affect network connectivity. When that happens, each UAV reasons in a decentralized, autonomous fashion to overcome the issues. As mentioned earlier, the key to taking into account, and hopefully compensating for, any unexpected circumstances is to actively employ, in the reasoning processes, realistic and up-to-date information about the communications state. 

The control loop used by each UAV is shown in Figure~\ref{fig:control-loop}a. In line with \cite{gl91,mt99,bar03}, the loop and the I/O functions are implemented procedurally, while the reasoning functions ($Goal\_Achieved$, $Unexpected\_Observations$, $Explain\_Observations$, $Compute\_Plan$) are implemented in ASP. The loop takes in input the mission goal and the mission plan, which potentially includes courses of actions for multiple UAVs. Functions $\mathrm{New\_Observations}$, $\mathrm{Next\_Action}$, $\mathrm{Tail}$, $\mathrm{Execute}$, $\mathrm{Record\_Execution}$ perform basic manipulations of data structures, and interface the agent with the  execution and perception layers. Functions $Next\_Action$ and $\mathrm{Tail}$ are assumed to be capable of identifying the portions of the mission plan that are relevant to the UAV executing the loop. The remaining functions occurring in the control loop implement the reasoning tasks. Central to the architecture is the maintenance of a history of past observations and actions executed by the agent. Such history is stored in variable $H$ and updated by the agent when it gathers observations about its environment and when it performs actions. It is important to note that variable $H$is local to the specific agent executing the loop, rather than shared among the UAVs (which would be highly unrealistic in a communication-constrained environment). Thus,  different agents will develop differing views of the history of the environment as execution unfolds. At a minimum, the difference will be due to the fact that agents cannot observe each other's actions directly, but only their consequences, and even those are affected by the partial observability of the environment.

Details on the control loop can be found in \cite{bg08}. With respect to that version of the loop, the control loop used in the present work does not allow for the selection of a new goal at run-time, but it extends the earlier control loop with the ability to deal with, and reason about, an externally-provided, multi-agent plan, and to reason about other agents' behavior. We do not expect run-time selection of goals to be difficult to embed in the control loop presented here, but doing so is out of the scope of the current phase of the project.   

\begin{figure}[htbp]
\begin{center}
\begin{tabbing}
iiii\=iiii\=iiii\=iiii\=iiii\=iiii\=iiii\=iiii\=iiii\kill
\textbf{Input:} \>\>\> $M$: mission plan;\\
\>\>\>$G$: mission goal;\\
\textbf{Vars:}\>\>\>$H$ : history; \\
\>\>\>$P$: current plan; \\
\\
\>$P := M$; \\
\>$H := \mathrm{New\_Observations}()$; \\
\>\textbf{while} $\neg \mathrm{Goal\_Achieved}(H,G)$ \textbf{do}\\
\>\>\textbf{if} $\mathrm{Unexpected\_Observations}(H)$ \textbf{then}\\
\>\>\>$H := \mathrm{Explain\_Observations}(H)$; \\
\>\>\>$P := \mathrm{Compute\_Plan}(G,H,P)$; \\
\>\>\textbf{end if}\\
\>\>$A := \mathrm{Next\_Action}(P)$; \\
\>\>$P := \mathrm{Tail}(P)$; \\
\>\>$\mathrm{Execute}(A)$; \\
\>\>$H := \mathrm{Record\_Execution}(H,A)$; \\
\>\>$H := H \cup \,\mathrm{New\_Observations}()$; \\
\>\textbf{loop}\\
\end{tabbing}
\end{center}
\caption{Agent Control Loop.}\label{fig:control-loop}
\end{figure}
\begin{figure}[htbp]
  \centering
  \begin{subfigure}[t]{\linewidth}%{.45\linewidth}
    \includegraphics[width=\textwidth]{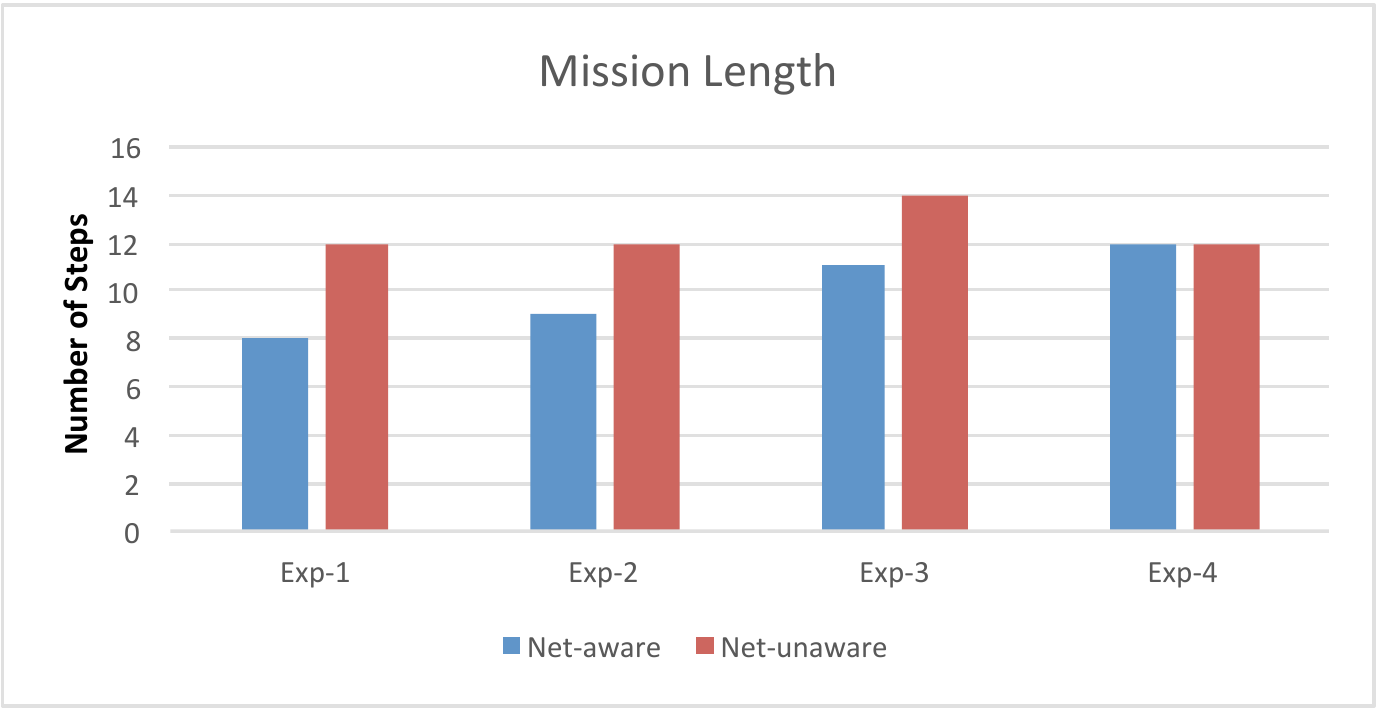}
    \subcaption{Length of the mission in time steps for the example instances.}\label{subfig:mission-length}
  \end{subfigure}%\ \ \ 
  \vspace{.4in}
  \\
  \begin{subfigure}[t]{\linewidth}%{.45\linewidth}
    \centering
    \includegraphics[width=\linewidth]{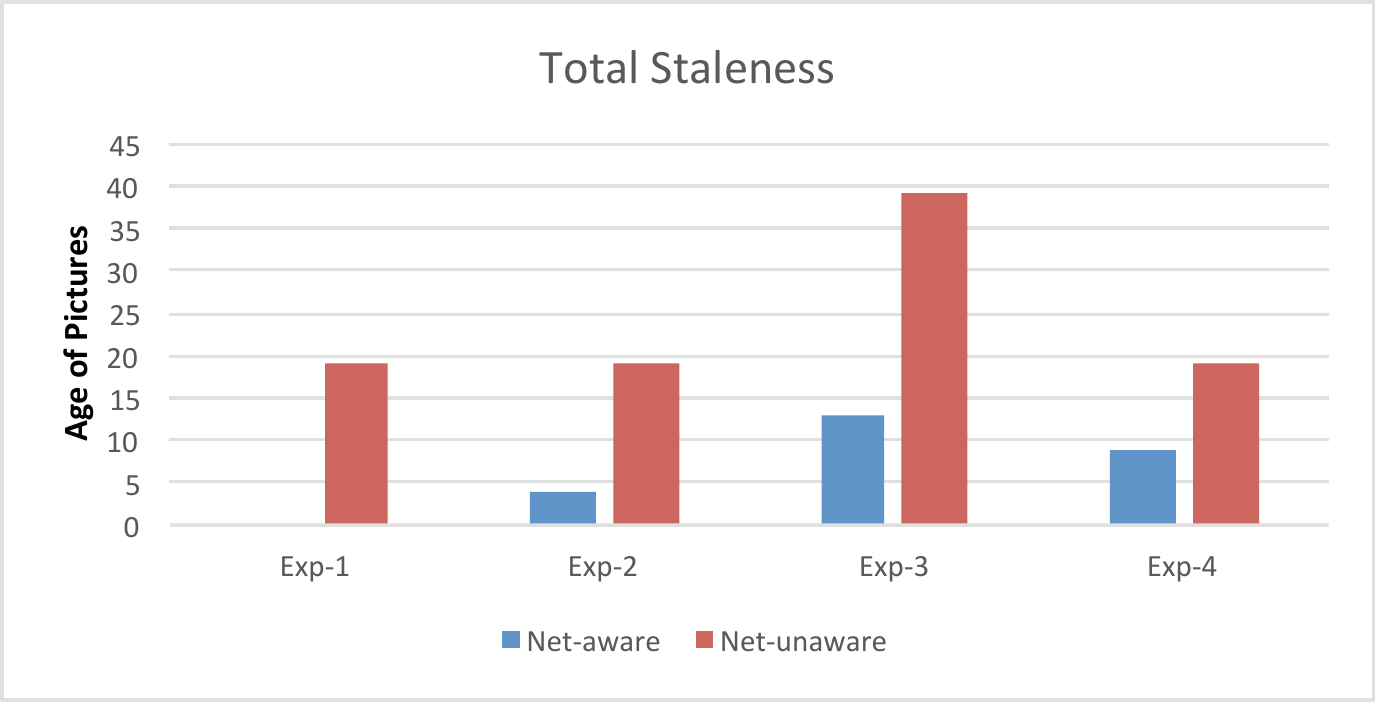}
    \subcaption{The total staleness of the image transfers.}\label{subfig:total-staleness}
  \end{subfigure}
\caption{Performance comparison.}\label{fig:performance}
\end{figure}

%The example scenario, while simplified, provides a suitable situation
%where a planner capable of incorporating network information in state
%evaluation. The planner is formally defined as such 

\begin{figure*}[htbp]
  \centering
  \begin{subfigure}[t]{0.42\linewidth}%{0.21\linewidth}
    \centering
    \includegraphics[width=\textwidth]{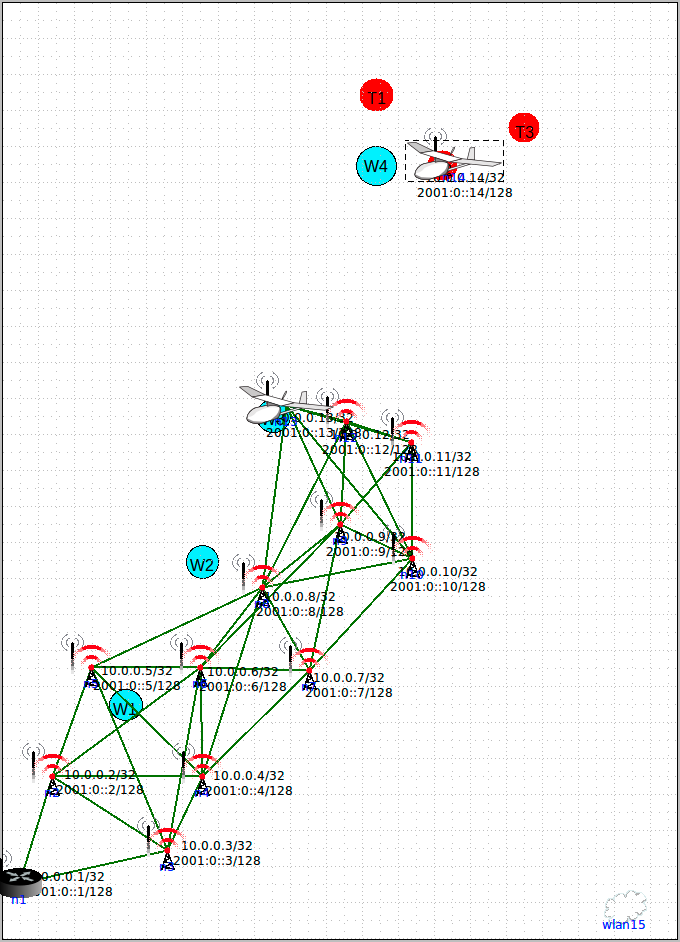}
    \subcaption{Step 5: $u_1$ is disconnected from home base.}\label{subfig:datamule}
  \end{subfigure}
  \quad
  \begin{subfigure}[t]{0.42\linewidth}%{0.21\linewidth}
    \centering
    \includegraphics[width=\textwidth]{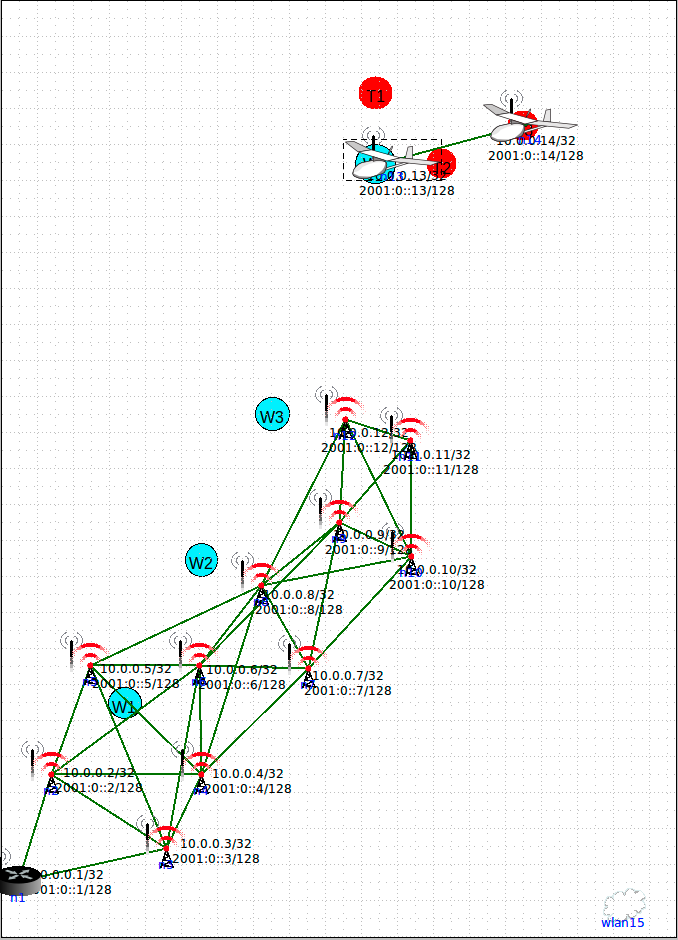}
    \subcaption{Step 6: $u_2$ connects with $u_1$
    and transfers images $t_2$ and $t_3$.}\label{subfig:datamule2} 
  \end{subfigure}\\
%  \quad
  \begin{subfigure}[t]{0.42\linewidth}%{0.21\linewidth}
    \centering
    \includegraphics[width=\linewidth]{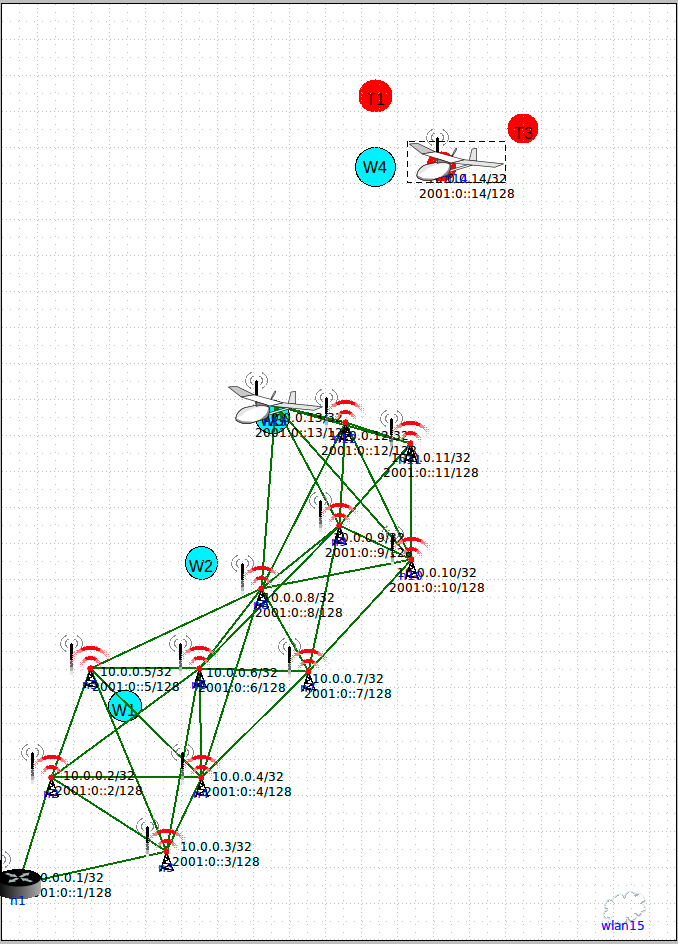}
    \subcaption{Step 7: $u_2$ reconnects
    with relays, transfers images to the home base.}\label{subfig:datamule3}
  \end{subfigure}
  \quad
  \begin{subfigure}[t]{0.42\linewidth}%{0.21\linewidth}
    \centering
    \includegraphics[width=\linewidth]{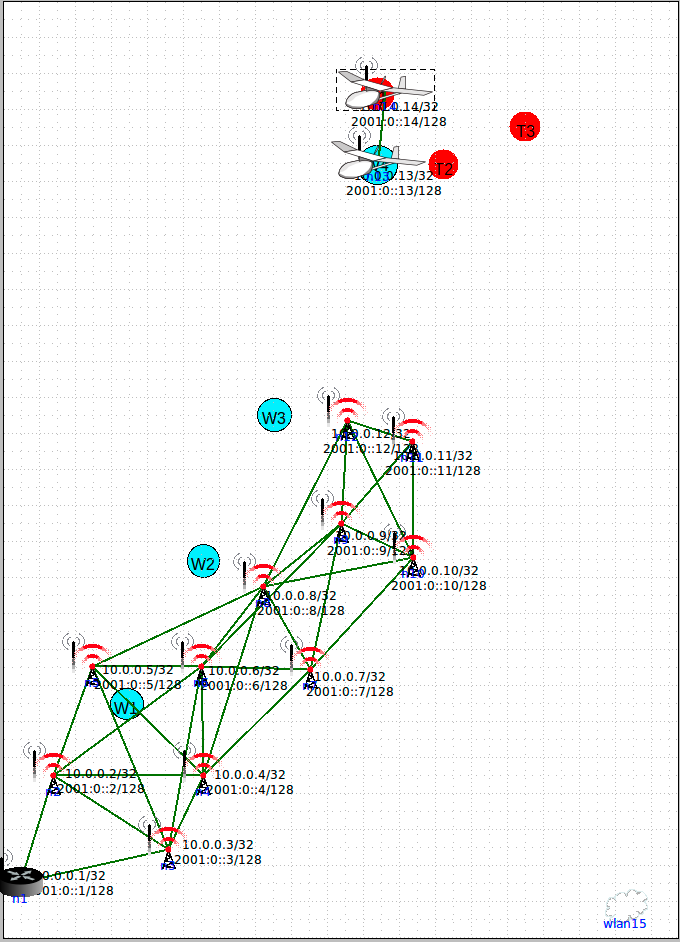}
    \subcaption{Step 8: $u_2$ reconnects
    with $u_1$ to relay images of $t_1$.}\label{subfig:datamule4}
  \end{subfigure}
  \caption{Example instance 1 illustrating ``data mule'' information
  relaying between $u_1$ and $u_2$. %The home base is the black node in the lower left corner and the targets are shown as red dots in the upper right corner. Relays form a mesh and extend the network.
  }\label{fig:inst2}
\end{figure*}

\subsection{Network-Aware Reasoning}
The major reasoning tasks (centralized mission planning, as well as anomaly detection, explanation and planning within each agent) are reduced to finding models of answer-set based
%\cite{gl91,bar03} 
formalizations of the corresponding problems.
Central to all the reasoning tasks is the ability to represent the evolution of the environment over time. Such evolution is conceptualized into a \emph{transition diagram} \cite{gl93}, a graph whose nodes correspond to states of the environment, and whose arcs describe state transitions due to the execution of actions. Let $\mathcal{F}$ be a collection of \emph{fluents}, expressions representing relevant properties of the domain that may change over time, and let $\mathcal{A}$ be a collection of \textit{actions}. A \emph{fluent literal} $l$\ is a fluent $f \in \mathcal{F}$ or its negation $\neg f$. A \emph{state }$\sigma$ is a complete and consistent set of fluent literals. 

The transition diagram is formalized in ASP\ by rules describing the direct effects of actions, their executability conditions, and their indirect effects (also called state constraints). The succession of moments in the evolution of the environment is characterized by discrete \emph{steps}, associated with non-negative integers. The fact that a certain  fluent $f$ is true at a step $s$ is encoded by an atom $h(f,s)$. If $f$ is false, this is expressed by $\neg h(f,s)$. The occurrence of an action $a \in \mathcal{A}$ at step $s$ is represented as $o(a,s)$.

The history of the environment is formalized in ASP\ by two types of statements: $obs(f,true,s)$ states that $f$ was observed to be true at step $s$ (respectively, $obs(f,false,s)$ states that $f$ was false); $hpd(a,s)$ states that $a$ was observed to occur at $s$. Because in the this paper other agents' actions are not observable, the latter expression is used  only to record an agent's own actions.

Objects in the UAV domain discussed in this paper are the home base, a set of fixed relays, a set of UAVs, a set of targets, and a set of waypoints. The waypoints are used to simplify the path-planning task, which we do not consider in the present work. The locations that the UAVs can occupy and travel to are the home base, the waypoints, and the locations of targets and fixed relays. The current location, $l$, of UAV $u$ is represented by a fluent $at(u,l)$. For each location, the collection of its neighbors is defined by relation $next(l,l')$. UAV motion is restricted to occur only from a location to a neighboring one. The direct effect of action $move(u,l)$, intuitively stating that UAV $u$ moves to location $l$, is described by the rule:
\[
\begin{array}{l}
h(at(U,L2),S+1) \hif \\
\aspindent o(move(U,L2),S), \\
\aspindent h(at(U,L1),S), \\
\aspindent next(L1,L2).
\end{array}
\]
The fact that two radio nodes are in radio contact is encoded by fluent $in\_contact(r_1,r_2)$.
The next two rules provide a recursive definition of the fluent, represented by means of state constraints:
\[
\begin{array}{l}
h(in\_contact(R1,R2),S) \hif \\
\aspindent R1 \neq R2, \\
\aspindent \neg h(down(R1),S), \ \neg h(down(R2),S), \\
\aspindent h(at(R1,L1),S), \ h(at(R2,L2),S), \\
\aspindent range(Rg), \\
\aspindent dist2(L1,L2,D), \ D \leq Rg^2. \\
\\
h(in\_contact(R1,R3),S) \hif \\
\aspindent R1 \neq R2,\ R2 \neq R3,\ R1 \neq R3, \\
\aspindent \neg h(down(R1),S),\ \neg h(down(R2),S). \\
\aspindent h(at(R1,L1),S),\ h(at(R2,L2),S), \\
\aspindent range(Rg), \\
\aspindent dist2(L1,L2,D),\ D \leq Rg^2,\\
\aspindent h(in\_contact(R2,R3),S),\\
\end{array}
\]
The first rule defines the base case of two radio nodes that are directly in range of each other. Relation $dist2(l_1,l_2,d)$ calculates the square  of the distance between two locations. Fluent $down(r)$ holds if radio $r$ is known to be out-of-order, and a suitable axiom (not shown) defines the closed-world assumption on it. In the formalization, $in\_contact(R1,R2)$ is a \emph{defined positive fluent}, i.e., a fluent whose truth value, in each state, is completely defined by the current value of other fluents, and is not subject to inertia. The formalization of $in\_contact(R1,R2)$ is thus completed by a rule capturing the closed-world assumption on it:
\[
\begin{array}{l}
\neg h(in\_contact(R1,R2),S) \hif \\
\aspindent R1 \neq R2,\\
\aspindent \lpnot h(in\_contact(R1,R2),S).
\end{array}
\]
Functions $\mathrm{Goal\_Achieved}$ and $\mathrm{Unexpected\_Observations}$, in Figure~\ref{fig:control-loop}a, respectively check if the goal has been achieved, and whether the history observed by the agent contains any unexpected observations. Following the definitions from \cite{gb02}, observations are unexpected if they contradict the agent's expectations about the corresponding state of the environment. This definition is captured by the \emph{reality-check axiom}, consisting of the constraints:
\[
\begin{array}{l}
\hif obs(F,true,S),\ \neg h(F,S). \\
\hif obs(F,false,S),\ h(F,S).
\end{array}
\]
Function $\mathrm{Explain\_Observations}$ uses a diagnostic process along the lines of \cite{gb02} to identify a set of exogenous actions (actions beyond the control of the agent that may occur unobserved), whose occurrence explains the observations. To deal with the complexities of reasoning in a dynamic, multi-agent domain, the present work extends the previous results on diagnosis by considering multiple types of exogenous actions, and preferences on the resulting explanations. The simplest type of exogenous action is $break(r)$, which occurs when radio node $r$ breaks. This action causes fluent $down(r)$ to become true. Actions of this kind may be used to explain unexpected observations about the lack of radio contact. However, the agent must also be able to cope with the limited observability of the position and motion of the other agents. This is accomplished by encoding commonsensical statements (encoding omitted) about the behavior of other agents, and about the factors that may affect it. The first such statement says that \emph{a UAV will normally perform
the mission plan, and will stop
performing actions when its portion of the mission plan is complete.} Notice that a mission plan is simply a sequence of actions. There is no need to include pre-conditions for the execution of the actions it contains, because those can be easily identified by each agent, at execution time, from the formalization of the domain.

The agent is allowed to hypothesize that \emph{a UAV  may have stopped executing the mission plan} (for example, if the UAV malfunctions or is destroyed).  \emph{Normally, the reasoning agent will expect a UAV that aborts execution to remain in its latest location.} In certain circumstances, however, a UAV may need to deviate completely from the mission plan. To accommodate for this situation, the agent may hypothesize that \emph{a UAV began behaving in an unpredictable way (from the agent's point of view) after aborting plan execution}. The following choice rule allows an agent to consider all of the possible explanations:
\[
\begin{array}{l}
\{\ hpd(break(R),S) , hpd(aborted(U,S)), \\
\ \ \ hpd(unpredictable(U,S)) \ \}.
\end{array}
\]
A constraint ensures that unpredictable behavior can be considered only if a UAV is believed to have aborted the plan. If that happens,  the following choice rule is used to consider all  possible courses of actions from the moment the UAV became unpredictable to the current time step.
%This makes it possible to explain the available observations about the UAV's location.
\[
\begin{array}{l}
\{ hpd(move(U,L),S') : S' \geq S : S' < currstep \} \hif \\
\aspindent hpd(unpredictable(U,S)).
\end{array}
\]In practice, such a thought process is important to enable coordination with other UAVs when communications between them are impossible, and to determine the side-effects of the inferred courses of actions and potentially take advantage of them (e.g., ``the UAV must have flown by target $t_3$. Hence, it is no longer necessary to take a picture of $t_3$''). A \emph{minimize} statement ensures that only cardinality-minimal diagnoses are found:
\[
\begin{array}{l}
\#minimize[ hpd(break(R),S), \\
\hspace*{.77in}hpd(aborted(U,S)), \\
\hspace*{.77in}hpd(unpredictable(U,S)) ].
\end{array}
\]

An additional effect of this statement is that the reasoning agent will prefer simpler explanations, which assume that a UAV aborted the execution of the mission plan and stopped, over those hypothesizing that the UAV engaged in an unpredictable course of actions.

Function $\mathrm{Compute\_Plan}$, as well as the mission planner, compute a new plan using a rather traditional approach, which relies on a choice rule for generation of candidate sequences of actions, constraints to ensure the goal is achieved, and \emph{minimize} statements to ensure  optimality of the plan with respect to the given metrics.

The next paragraphs outline two experiments, in increasing order of sophistication, which demonstrate the features of our approach, including non-trivial emerging interactions between the UAVs and the ability to work around unexpected problems autonomously.

% \emph{(insert instances from slides)} slide 8. describe the 3 instances as in figure inst1_00_init

% describe inst 2 and 3 and provide key figures for instance 2 and 3.
%Instance2:(dont' mention waypoints)

\noindent
\textbf{Example Instance 1.} Consider the environment shown in in Figure~\ref{fig:inst2}. Two UAVs, $u_1$ and $u_2$ are initially located at the home base in the lower left corner. The home base, relays and targets are positioned as shown in the figure, and the radio range is set to $7$ grid units. 

The mission planner finds a plan in which the UAVs begin by traveling toward the targets. While $u_1$ visits the first two targets, $u_2$ positions itself so as to be in radio contact with $u_1$ (Figures~\ref{subfig:datamule} and~\ref{subfig:datamule2}). Upon receipt of the pictures, $u_2$ moves to within range of the relays to transmit the pictures to the home base (Figure~\ref{subfig:datamule3}). At the same time, $u_1$ flies toward the final target. UAV $u_2$, after transmitting pictures to home base, moves to re-establish radio contact with $u_1$ and to receive the picture of $t_3$ (Figure~\ref{subfig:datamule4}). Finally, $u_2$ moves within range of the relays to transmit picture of $t_3$ to the home base.

Remarkably, in this problem instance the plan establishes $u_2$ as a "data mule"  in order to cope with the network limits. The "data mule" behavior is well-known in sensor network applications~\cite{Shah2003215,sensorsdatamule2005}; however, no description of such behavior is included in our planner. Rather, the behavior emerges as a result of the reasoning process. The data-mule behavior is adopted by the planner because it optimizes the evaluation metrics (mission length and total staleness).

\begin{figure*}[ht]
  \centering
  \begin{subfigure}[t]{0.30\linewidth}%{0.19\linewidth}
    \centering
    \includegraphics[width=\textwidth]{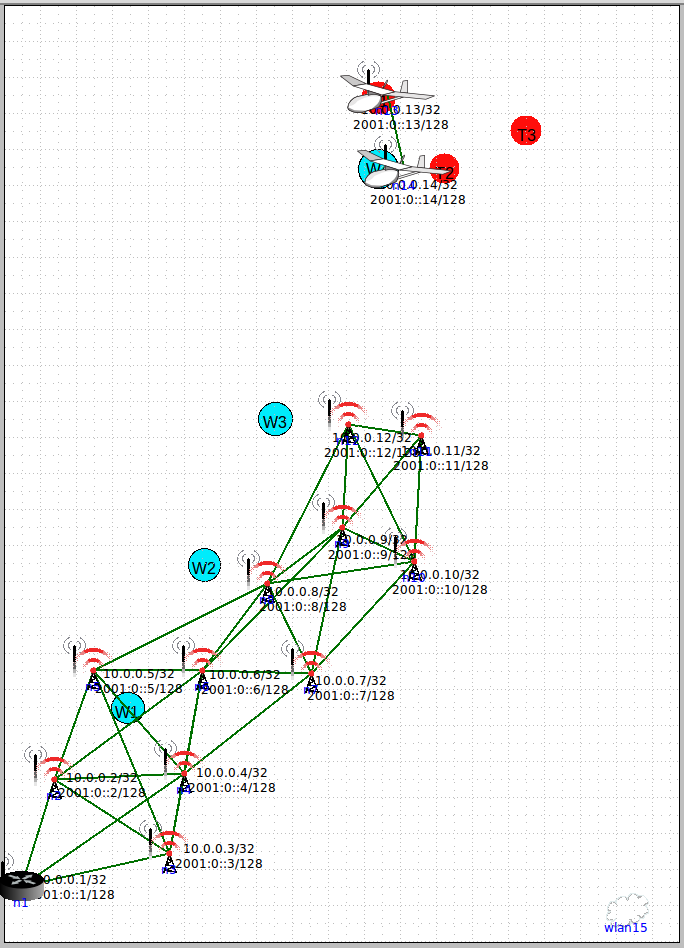}
    \subcaption{Step 5: $u_1$ is transmitting images to $u_2$.}\label{subfig:relays}
  \end{subfigure}
  \quad
  \begin{subfigure}[t]{0.30\linewidth}%{0.19\linewidth}
    \centering
    \includegraphics[width=\textwidth]{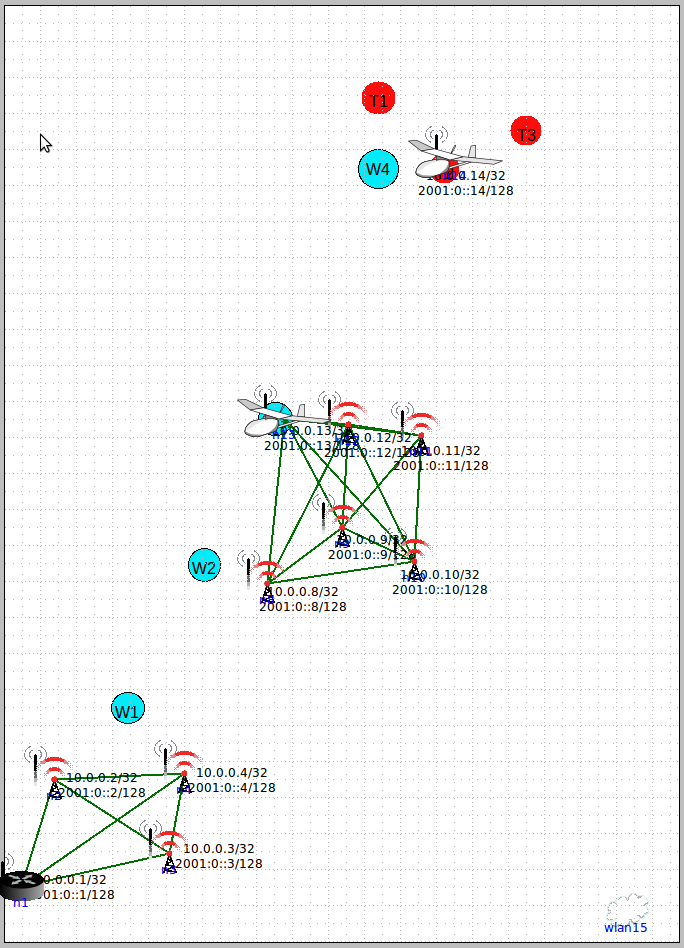}
    \subcaption{Step 6: $u_2$ moves toward relays. Relay nodes 5, 6, and 7 have failed.}\label{subfig:relays2} 
  \end{subfigure}
  \quad
  \begin{subfigure}[t]{0.30\linewidth}%{0.19\linewidth}
    \centering
    \includegraphics[width=\linewidth]{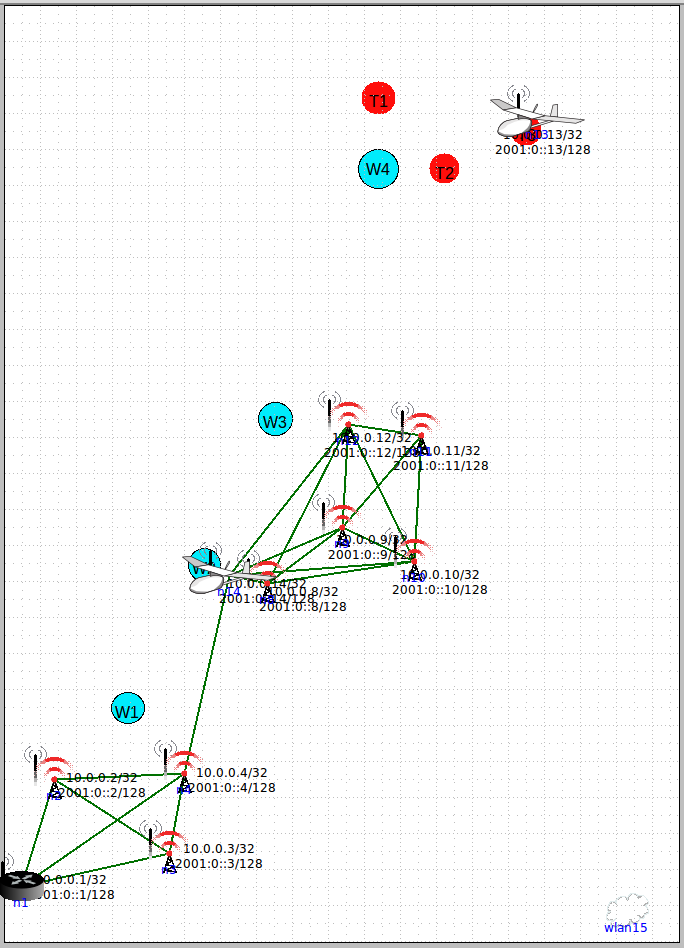}
    \subcaption{Step 7: $u_2$ re-plans and moves closer to home base.}\label{subfig:relays3}
  \end{subfigure}
  %\quad
  \\
  \begin{subfigure}[t]{0.30\linewidth}%{0.19\linewidth}
    \centering
    \includegraphics[width=\textwidth]{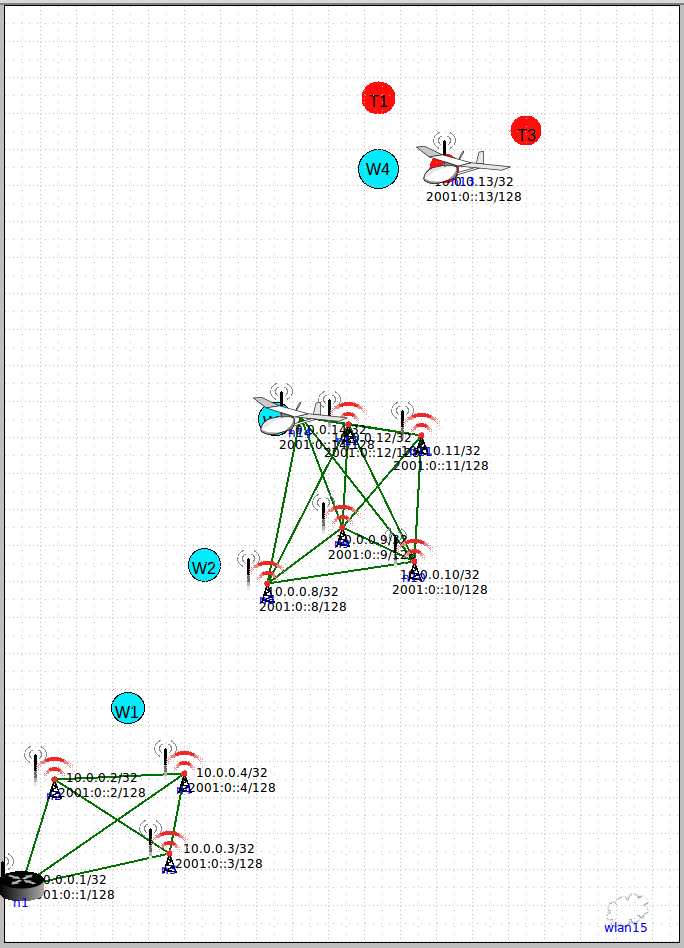}
    \subcaption{Step 8: $u_2$ moves toward $u_1$.}\label{subfig:relays4}
  \end{subfigure}
  \quad
  \begin{subfigure}[t]{0.30\linewidth}%{0.19\linewidth}
    \centering
    \includegraphics[width=\textwidth]{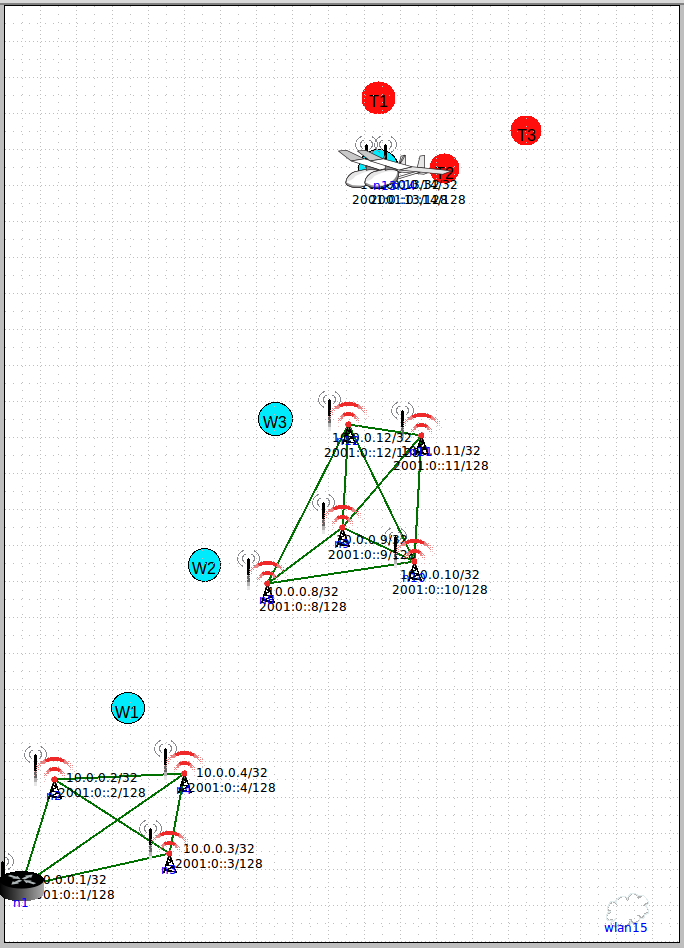}
    \subcaption{Step 9: $u_2$ and $u_1$ reconnect and move back toward home base.}\label{subfig:relays5}
  \end{subfigure}
  %\quad
  \caption{Example instance 2 illustrates re-planning after relay node failure between steps 5 and 6 forcing the UAVs to re-plan.}\label{fig:inst3}
\end{figure*}

\noindent
\textbf{Example Instance 2.} Now consider a more challenging and realistic example (Figure~\ref{fig:inst3}), in which the UAVs must cope with unexpected events occurring during mission execution. Environment and mission goals are as above.

The mission planner produces the same plan described earlier\footnote{The careful reader may notice from the figures that the trajectory used to visit the targets is the mirror image of the one from the previous example. The corresponding plans are equivalent from the point of view of all the metrics, and the specific selection of one over the other is due to randomization used in the search process.}, in which $u_2$ acts as a ``data mule.''
The execution of the plan begins as expected, with $u_1$ reaching the area of the targets and $u_2$ staying in radio contact with it in order to receive the pictures of the first two targets (Figure~\ref{subfig:relays}).
When $u_2$ flies back to re-connect with the relays, however, it observes (``Observe'' step of the architecture from Figure~\ref{fig:arch}b) that the home base is unexpectedly not in radio contact. Hence, $u_2$ uses the available observations to determine plausible causes (``Explain'' step of the architecture). In this instance, $u_2$ observes that relays $r_5$, $r_6$, $r_7$ and all the network nodes South of them are not reachable via the network. Based on knowledge of the layout of the network, $u_2$ determines that the simplest plausible explanation is that those three relays must have stopped working while $u_2$ was out of radio contact (e.g., started malfunctioning or have been destroyed).\footnote{As shown in Figure~\ref{subfig:relays2} this is indeed the case in our experimental set-up, although it need not be. Our architecture is capable of operating under the \emph{assumption} that its hypotheses are correct, and later re-evaluate the situation based on further observations, and correct its hypotheses and re-plan if needed.}
Next, $u_2$ replans (``Local Planner'' step of the architecture). \emph{The plan is created based on the assumption that $u_1$ will continue executing the mission plan. This assumption can be later withdrawn if observations prove it false.} Following the new plan, $u_2$ moves further South towards the home base (Figure~\ref{subfig:relays3}). Simultaneously, $u_1$ continues with the execution of the mission plan, unaware that the connectivity has changed and that $u_2$ has deviated from the mission plan.
After successfully relaying the pictures to the home base, $u_2$ moves back towards $u_1$.  UAV $u_1$, on the other hand, reaches the expected rendezvous point, and observes that $u_2$ is not where expected (Figure~\ref{subfig:relays4}).  UAV $u_1$ does not know the actual position of $u_2$, but its absence is evidence that $u_2$ must have deviated from the  plan at some  point in time. Thus, $u_1$'s must now replan. Not knowing $u_2$'s state, $u_1$'s plan is to fly South to relay the missing picture to the home base on its own. This plan still does not deal with the unavailability of $r_5$, $r_6$, $r_7$, since $u_1$ has not yet had a chance to get in radio contact with the relays and observe the current network connectivity state.
The two UAVs continue with the execution of their new plans and eventually meet, unexpectedly for both (Figure~\ref{subfig:relays5}). At that point, they automatically share the final picture. Both now determine that the mission can be completed by flying South past the failed relays, and execute the corresponding actions.

%describe instance3: mission plan constructed as before however unknown to the mission planner, relays go down. slide 30. relays go down.
%during plan execution, a set of relays "go down" disconnecting the network from home basebefore $u_2$ has a chance to relay pictures  (as shown in figure (slide 39)). (ignore "note:" maybe) 

\noindent
\textbf{Experimental Comparison.} As mentioned earlier, we believe that our network-aware approach to reasoning provides advantages over the state-of-the-art techniques that either disregard the network, or assume perfect communications.
Figure~\ref{fig:performance}b provides an overview of a quantitative experimental demonstration of such advantages. The figure compares our approach with the one in which the network is disregarded, in terms of mission length and total staleness.\footnote{For simplicity we measure mission length and staleness in time steps, but it is not difficult to add action durations.} The optimistic approach is not considered, because its brittleness makes it not viable for actual applications. The comparison includes the two example instances discussed earlier (labeled Exp-2 and Exp-4). Of the other two experiments, Exp-1 is a variant of Exp-2 that can be solved with the data-mule~ in a static position, while Exp-3 is a variant of Exp-2 with $5$ targets. As can be seen, the network-aware approach is always superior. In Exp-1, the UAV acting as a data-mule  extends the range of the network so that all the pictures are instantly relayed to the home base, reducing total staleness to $0$. In Exp-4, it is worth stressing that the network, which the UAVs rely upon when using our approach, suddenly fails. One would expect the network-\emph{unaware} approach to have an advantage under these circumstances, but, as demonstrated by the experimental results, our approach still achieves a lower total staleness of the pictures thanks to its ability to identify the network issues and to work around them.

From a practical perspective, the execution times of the various reasoning tasks have been extremely satisfactory, taking only fractions of a second on a modern desktop computer running the {\sc clasp} solver \cite{gks09}, even in the most challenging cases.   
\if 0
\begin{figure}[t]
  \centering
  \begin{subfigure}[t]{.4\linewidth}
    \centering
    \includegraphics[width=\textwidth]{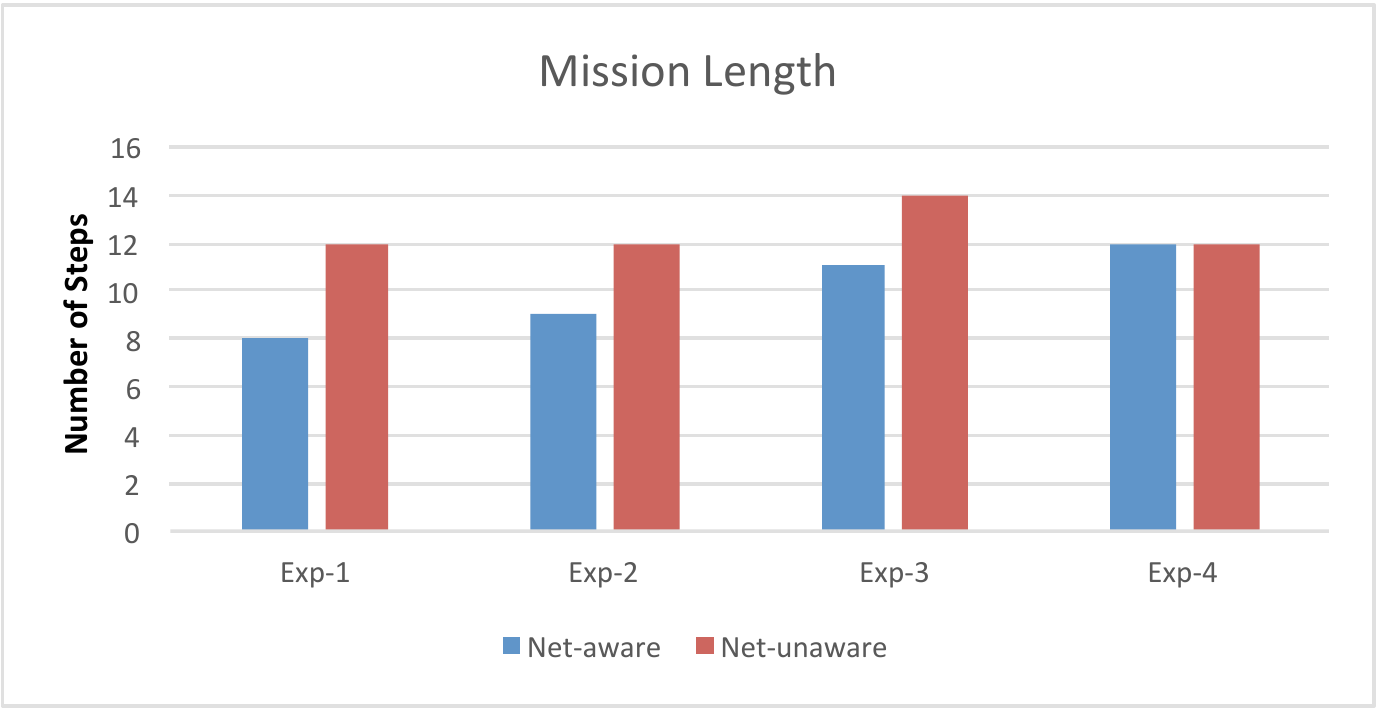}
    \subcaption{Length of the mission in time steps for the example instances.}\label{subfig:mission-length} 
  \end{subfigure}
  \begin{subfigure}[t]{.4\linewidth}
    \centering
    \includegraphics[width=\linewidth]{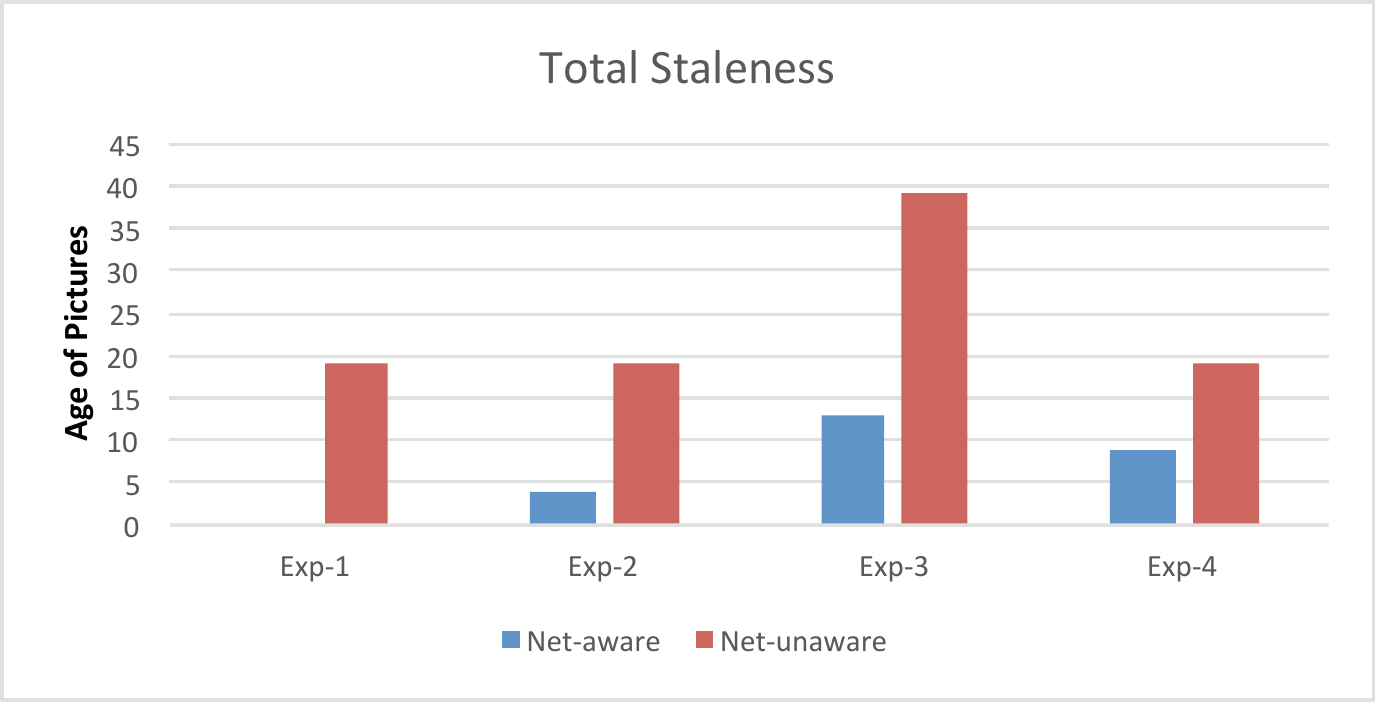}
    \subcaption{The total staleness of the image transfers.}\label{subfig:total-staleness}
  \end{subfigure}
  \caption{Performance comparison.}\label{fig:performance}
\end{figure}
\fi
%slide 40: $u_2$ replan based on avail info, decides to move closer to homebase to transmit images.
%at the same time $u_1$ is unaware of situation and continues visiting targets. (slide42) $u_1$ is expecting $u_2$ to return. whe $u_2$ is not available to receive $u_1$ pictures, $u_1$ decides that $u_2$ has abandoned mission plan for unknown reasons, $u_1$ replans and moves toward homebase to transmit images.

%slide42 uavs unexpectedly meet after replanning, share pictures near the targets. and both move toward home base to relay pictures to complete the mission. $u_1$ still unaware network is down. (add picture)

\section{Simulation and Experimental Setup}

The simulation for the experimental component of this work was built using the Common Open Research Emulator (CORE)~\cite{ahrenholz10core}. CORE is a real-time network emulator that allows users to create lightweight virtual nodes with full-fledged network communications stack. CORE virtual nodes can run unmodified Linux applications in real-time. The CORE GUI incorporates a basic range-based model to emulate networks typical in mobile ad-hoc network (MANET) environments. CORE provides an interface for creating complex network topologies, node mobility in an environment, and access to the lower-level network conditions, e.g., network connectivity.
Using CORE as a real-time simulation environment allows agents, represented as CORE nodes, to execute mission plans in realistic radio environments. For this work, CORE router nodes represent the home base, relays, and UAVs. The nodes are interconnected via an ad-hoc wireless network. As the UAVs move in the environment, CORE updates the connectivity between other UAVs and relays based on the range dictated by the built-in wireless model. The radio network model has limited range and bandwidth capacity. Each node runs the Optimized Link-State Routing protocol (OLSR)~\cite{jacquet2001optimized}, a unicast MANET routing algorithm, which maintains the routing tables across the nodes. The routing table makes it possible to determine if a UAV can exchange information with other radio nodes at any given moment. Using CORE allows us to account for realistic communications in ways not possible with multi-agent simulators such as AgentFly~\cite{sislak2012agentfly}.

%During network-aware planning simulations, the 
%In the simulation, the wireless network bandwidth is limited to 512kbit/s. Relay 10 provides a constant video stream to all video nodes consuming 100kbits/s across the network. In a plan-unaware network, at plan step 5, $u_2$ takes on average 72 seconds to transmit the file to homebase. In the plan-aware network, with the video stream not running, $u_2$ takes on average 24 seconds to transmit the file.

\section{Conclusion and Future Work} 

This paper discussed a novel application of an ASP-based intelligent agent architecture to the problem of UAV coordination.
% The problems of planning a mission and coordinating the communications activities during the mission are tightly coupled.  
The UAV scenarios considered in this paper are bound to be increasingly common as more levels autonomy are required to create large-scale systems. Prior work on distributed coordination and planning has mostly overlooked or simplified communications dynamics, at best treating communications as a resource or other planning constraint. 

Our work demonstrates the reliability and performance gains deriving from network-aware reasoning. In our experimental evaluation, our approach yielded a reduction in mission length of up to $30\%$ and in total staleness between $50\%$ and $100\%$. We expect that, in more complex scenarios, the advantage of a realistic networking model will be even more evident. In our experiments, execution time was always satisfactory, and we believe that several techniques from the state-of-the-art can be applied to curb the increase in execution time as the scenarios become more complex. For the future, we intend to extend the mission-aware networking layer with advanced reasoning capabilities, integrate network-aware reasoning and mission-aware networking tightly, and execute experiments demonstrating the advantages of such a tight integration.

%Before our approach can reach the deployment level, methods will have to be developed enabling the propagation through the network of the plans generated by the UAVs, allowing for UAVs to coordinate their re-planning efforts whenever they are in radio contact, and plan execution and monitoring in network-centric environments.

% (collaborative replanning as future work)

%Plan-aware networking
%Need: 
%- Common language describing the available resources for communications between agents. network ontology
%- 

%\newpage
\bibliographystyle{aaai}
\bibliography{maranets-paper,biblio-mb-mod}

\begin{thebibliography}{}

\bibitem[\protect\citeauthoryear{Ahrenholz}{2010}]{ahrenholz10core}
Ahrenholz, J.
\newblock 2010.
\newblock Comparison of {CORE} network emulation platforms.
\newblock In {\em IEEE Military Communications Conf.}

\bibitem[\protect\citeauthoryear{Balduccini and Gelfond}{2003}]{gb02}
Balduccini, M., and Gelfond, M.
\newblock 2003.
\newblock {D}iagnostic reasoning with {A}-{P}rolog.
\newblock {\em Journal of Theory and Practice of Logic Programming (TPLP)}
  3(4--5):425--461.

\bibitem[\protect\citeauthoryear{Balduccini and Gelfond}{2008}]{bg08}
Balduccini, M., and Gelfond, M.
\newblock 2008.
\newblock {T}he {A}{A}{A} {A}rchitecture: {A}n {O}verview.
\newblock In {\em AAAI Spring Symp.: Architectures for Intelligent Theory-Based
  Agents}.

\bibitem[\protect\citeauthoryear{Baral and Gelfond}{2000}]{bg00}
Baral, C., and Gelfond, M.
\newblock 2000.
\newblock {R}easoning {A}gents {I}n {D}ynamic {D}omains.
\newblock In {\em Workshop on Logic-Based Artificial Intelligence},  257--279.
\newblock Kluwer Academic Publishers.

\bibitem[\protect\citeauthoryear{Baral}{2003}]{bar03}
Baral, C.
\newblock 2003.
\newblock {\em {K}nowledge {R}epresentation, {R}easoning, and {D}eclarative
  {P}roblem {S}olving}.
\newblock Cambridge University Press.

\bibitem[\protect\citeauthoryear{Blount, Gelfond, and Balduccini}{2014}]{bgb14}
Blount, J.; Gelfond, M.; and Balduccini, M.
\newblock 2014.
\newblock {T}owards a {T}heory of {I}ntentional {A}gents.
\newblock In {\em Knowledge Representation and Reasoning in Robotics}, AAAI
  Spring Symp. Series.

\bibitem[\protect\citeauthoryear{David~Sislak and
  Pechoucek}{2012}]{sislak2012agentfly}
David~Sislak, Premysl~Volf, S.~K., and Pechoucek, M.
\newblock 2012.
\newblock {\em AgentFly: Scalable, High-Fidelity Framework for Simulation,
  Planning and Collision Avoidance of Multiple UAVs}.
\newblock Wiley Inc.
\newblock chapter~9,  235--264.

\bibitem[\protect\citeauthoryear{Gebser, Kaufmann, and Schaub}{2009}]{gks09}
Gebser, M.; Kaufmann, B.; and Schaub, T.
\newblock 2009.
\newblock {T}he {C}onflict-{D}riven {A}nswer {S}et {S}olver clasp: {P}rogress
  {R}eport.
\newblock In {\em Logic Programming and Nonmonotonic Reasoning}.

\bibitem[\protect\citeauthoryear{Gelfond and Lifschitz}{1991}]{gl91}
Gelfond, M., and Lifschitz, V.
\newblock 1991.
\newblock {C}lassical {N}egation in {L}ogic {P}rograms and {D}isjunctive
  {D}atabases.
\newblock {\em New Generation Computing} 9:365--385.

\bibitem[\protect\citeauthoryear{Gelfond and Lifschitz}{1993}]{gl93}
Gelfond, M., and Lifschitz, V.
\newblock 1993.
\newblock {R}epresenting {A}ction and {C}hange by {L}ogic {P}rograms.
\newblock {\em Journal of Logic Programming} 17(2--4):301--321.

\bibitem[\protect\citeauthoryear{Jacquet \bgroup et al\mbox.\egroup
  }{2001}]{jacquet2001optimized}
Jacquet, P.; Muhlethaler, P.; Clausen, T.; Laouiti, A.; Qayyum, A.; and
  Viennot, L.
\newblock 2001.
\newblock Optimized link state routing protocol for ad hoc networks.
\newblock In {\em {IEEE} {INMIC}: Technology for the 21st Century}.

\bibitem[\protect\citeauthoryear{Jea, Somasundara, and
  Srivastava}{2005}]{sensorsdatamule2005}
Jea, D.; Somasundara, A.; and Srivastava, M.
\newblock 2005.
\newblock Multiple controlled mobile elements (data mules) for data collection
  in sensor networks.
\newblock {\em Distr. Computing in Sensor Sys.}

\bibitem[\protect\citeauthoryear{Kopeikin \bgroup et al\mbox.\egroup
  }{2013}]{kpj13}
Kopeikin, A.~N.; Ponda, S.~S.; Johnson, L.~B.; and How, J.~P.
\newblock 2013.
\newblock {D}ynamic {M}ission {P}lanning for {C}ommunication {C}ontrol in
  {M}ultiple {U}nmanned {A}ircraft {T}eams.
\newblock {\em Unmanned Systems} 1(1):41--58.

\bibitem[\protect\citeauthoryear{Marek and Truszczynski}{1999}]{mt99}
Marek, V.~W., and Truszczynski, M.
\newblock 1999.
\newblock {\em {T}he {L}ogic {P}rogramming {P}aradigm: a 25-{Y}ear
  {P}erspective}.
\newblock Springer Verlag, Berlin.
\newblock chapter {S}table {M}odels and an {A}lternative {L}ogic {P}rogramming
  {P}aradigm,  375--398.

\bibitem[\protect\citeauthoryear{McCarthy}{1998}]{mcc98}
McCarthy, J.
\newblock 1998.
\newblock {E}laboration {T}olerance.

\bibitem[\protect\citeauthoryear{Niemel{\"a} and Simons}{2000}]{ns00}
Niemel{\"a}, I., and Simons, P.
\newblock 2000.
\newblock {\em {L}ogic-{B}ased {A}rtificial {I}ntelligence}.
\newblock Kluwer Academic Publishers.
\newblock chapter {E}xtending the {S}models {S}ystem with {C}ardinality and
  {W}eight {C}onstraints.

\bibitem[\protect\citeauthoryear{Pynadath and Tambe}{2002}]{pt02}
Pynadath, D.~V., and Tambe, M.
\newblock 2002.
\newblock {T}he {C}ommunicative {M}ultiagent {T}eam {D}ecision {P}roblem:
  {A}nalyzing {T}eamwork {T}heories and {M}odels.
\newblock {\em JAIR} 16:389--423.

\bibitem[\protect\citeauthoryear{Rao and Georgeff}{1991}]{rg91}
Rao, A.~S., and Georgeff, M.~P.
\newblock 1991.
\newblock {M}odeling {R}ational {A}gents within a {B}{D}{I}-{A}rchitecture.
\newblock In {\em Proc. of the Int'l Conf. on Principles of Knowledge
  Representation and Reasoning}.

\bibitem[\protect\citeauthoryear{Shah \bgroup et al\mbox.\egroup
  }{2003}]{Shah2003215}
Shah, R.~C.; Roy, S.; Jain, S.; and Brunette, W.
\newblock 2003.
\newblock Data {MULE}s: modeling and analysis of a three-tier architecture for
  sparse sensor networks.
\newblock {\em Ad Hoc Networks} 1(2-–3).

\bibitem[\protect\citeauthoryear{Usbeck, Cleveland, and
  Regli}{2012}]{usbeckCR12}
Usbeck, K.; Cleveland, J.; and Regli, W.~C.
\newblock 2012.
\newblock Network-centric ied detection planning.
\newblock {\em IJIDSS} 5(1):44--74.

\bibitem[\protect\citeauthoryear{Wooldridge}{2000}]{woo00}
Wooldridge, M.
\newblock 2000.
\newblock {\em {R}easoning about {R}ational {A}gents}.
\newblock MIT Press.

\end{thebibliography}

\end{document}